# Face Identification from Manipulated Facial Images using SIFT

H. R. Chennamma
Dept. of Studies in Computer Science
University of Mysore
Mysore, INDIA
anusha_hr@rediffmail.com

Lalitha Rangarajan
Dept. of Studies in Computer Science
University of Mysore
Mysore, INDIA
lali85arun@yahoo.co.in

Veerabhadrappa
Dept. of Studies in Computer Science
University of Mysore
Mysore, INDIA
veerabadrappa@yahoo.com

*Abstract—* Editing on digital images is ubiquitous. Identification of deliberately modified facial images is a new challenge for face identification system. In this paper, we address the problem of identification of a face or person from heavily altered facial images. In this face identification problem, the input to the system is a manipulated or transformed face image and the system reports back the determined identity from a database of known individuals. Such a system can be useful in mugshot identification in which mugshot database contains two views (frontal and profile) of each criminal. We considered only frontal view from the available database for face identification and the query image is a manipulated face generated by face transformation software tool available online. We propose SIFT features for efficient face identification in this scenario. Further comparative analysis has been given with well known eigenface approach. Experiments have been conducted with real case images to evaluate the performance of both methods.

## I. INTRODUCTION

Photo editing software tools are becoming more sophisticated and user friendly day by day. Face is an important biometric trait for the identification of a person. Forensic investigation and law enforcement is one of the major applications of face recognition problem. Rigorous research has been carried out so far for recognition of faces by considering different viewpoints, illuminations, facial expressions, occlusions etc. Changing appearance to hide the identity of a person is very common. Some examples are shown in Fig. 1. In which original face images are modified by altering almost all facial features like eyes, ears, nose, hair style, mouth, shape of the face. etc. Such identity modifications are simulated using face transformation software tool. In this paper, we deal with the problem of face identification from altered facial images.

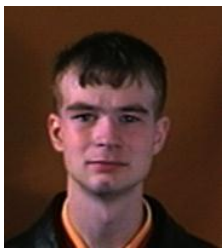 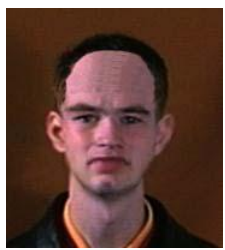

(a)   (b)

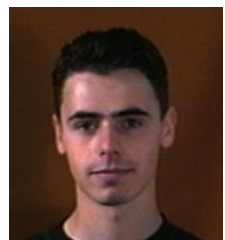 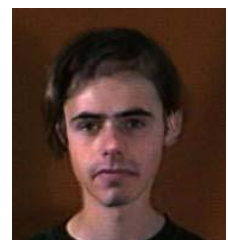

(c)   (d)

Figure 1. (a) & (c) are original images, (b) & (d) are modified images

Modifications to the face image are not a well defined notion and it is always depending on the purpose of usage. For a magazine cover or posters, skin softening and some local editing may be required. Change of complete appearance of the face may be necessary to misguide the face identification system or agency. In our face identification problem, the query face image to the system is suspected to be a manipulated face image and the system reports back the identified person from a face database of known individuals. In this research work, we concentrate only on mugshot identification. Mugshots consists of two views (frontal and profile) of each criminal. Mugshots are downloaded from **http://www.thesmokinggun.com/mugshots**. We considered only frontal view in our mugshot identification system. Image editing software tool like Adobe Photoshop is commonly used to perform alterations on digital images. For instance a skilled person can create old age face or childhood face from young adult face. Such a job can also be done by various software tools available on the web and such a process is called as face transformation. The aim of this research is to measure similarity between query (manipulated or transformed) face image with all the face images in database and retrieve the image which has got highest similarity i.e. nearest neighbour. Since query is created from one of the database image (source or original image), the system should assign highest rank to the right source image and retrieve it. Query image is created using the face transformation tool implemented by the University of St. Andrews available in **http://morph.cs.st-andrews.ac.uk.** Two well known face recognition techniques Scale Invariant Feature Transform (SIFT) and Eigenfaces are evaluated and their performances compared.

## II. RELATED WORK

As far as our knowledge, this is the first attempt for mugshot identification from modified face images. Here we review related work on face recognition problem that deal with different form of face representatives and we also review prior work that evaluated the robustness of SIFT features for face recognition.

Robert et. al [1] have presented a theory and practical computations for automatically matching a police artist sketch to a set of true photographs. This method locates facial features in both the sketch as well as the set of photograph images. Then, the sketch is photometrically standardized to facilitate comparison with a photo and then both the sketch and the photos are geometrically standardized. Finally, for matching, eigenanalysis is employed.

Xiaogang Wang et. al [2] have proposed a novel face photo-sketch synthesis and recognition method using a multi scale Markov Random Fields (MRF) model. To synthesize sketch/photo images, the face region is divided into overlapping patches for learning. From a training set which contains photo-sketch pairs, the joint photo-sketch model is learnt at multiple scales using a multiscale MRF model. By transforming a face photo to a sketch (or transforming a sketch to a photo), the difference between photos and sketches is significantly reduced, thus allowing effective matching between the two in face sketch recognition.

Wolfgang Konen [3] has compared facial line drawings with gray-level images using a software tool called PHANTOMAS. Yongsheng et. al [4] have presented a methodology for facial expression recognition from a single static using line-based caricature. The proposed approach uses structural and geometrical features of a user sketched expression model to match the Line Edge Map (LEM) descriptor of an input face image. A disparity measure that is robust to expression variations is defined.

Rich Singh et. al [5] have presented a novel age transformation algorithm to handle the challenge of facial aging in face recognition. The proposed algorithm registers the gallery and probe face images in polar coordinate domain and minimizes the variation in facial features caused due to aging. The efficiency of the proposed age transformation algorithm is validated using 2D log polar Gabor based face recognition algorithm on a face database that comprises of face images with large age progression.

Mohamed Aly [6] used SIFT features for general face recognition problem. He compared SIFT with Eigen faces and Fisher faces then reported the superiority of SIFT features for face recognition. Han Yanbin et. al [7] have extracted face features by using SIFT. Then, face recognition is conducted by comparing real extracted features with training sets. They experimented with ORL face database and reported recognition rate for SIFT, PCA, ICA and FLD as 96.3%, 92.5%, 91.6% and 92.8% respectively.

## III. BACKGROUND

### A. SIFT

Lowe [8] invented robust image features called Scale Invariant Feature Transform which are invariant to scale, rotation, affine transformations, noise, occlusions and are highly distinctive. Detection and representation of SIFT features consist of four major stages: (1) scale-space peak selection; (2) keypoint localization; (3) orientation assignment; (4) keypoint descriptor. In the first stage, potential interest points are identified by scanning the image over location and scale. This is implemented efficiently by constructing a Gaussian pyramid and searching for local peaks (termed keypoints) in a series of Difference-of-Gaussian (DoG) images. In the second stage, candidate keypoints are localized to sub-pixel accuracy and eliminated if found to be unstable. Stage 3 identifies the dominant orientations for each keypoint based on its location. The final stage builds a local image descriptor for each keypoint, based upon the image gradients in its local neighbourhood. Every feature is a vector of dimension 128 distinctively identifying the neighbourhood around the key point.

### B. PCA

Eigenfaces are based on the dimensionality reduction approach of Principal Component Analysis (PCA) [9]. The basic idea is to treat each image as a vector in a high dimensional space. Then PCA is applied to the set of images to produce a new reduced subspace that captures most of the variability between the input images. The Principal Component Vector (eigenvectors of the sample covariance matrix) is called the Eigenface. Every input image can be represented as a linear combination of these eigenfaces by projecting the image onto the new eigenfaces space. Then we can perform the identification process by matching in this reduced space. An input image is transformed into the eigenspace and the nearest face is identified using a nearest neighbor approach. Euclidean distance is used to match the input image against all images in the database.

## IV. APPROACH

Matlab is used to implement eigenfaces. Eigenfaces are computed for each face in the database and the eigenface of the query face is compared with all faces in the database. Comparison is done by computing Euclidean distance between two eigenfaces. Nearest neighbour of the query is retrieved which has got minimum distance.

The code for extracting SIFT features is available in Lowe's [8] website. The SIFT features are extracted from all faces in the database. Then given a new face image, the features extracted from that face are compared against the features from each face in the database. A feature is considered as matched with another feature when the distance to that feature is less than a specific fraction of the distance to the next nearest feature. Further spatial topology is verified by Angle-Line Ratio (ALR) statistics [10] among the matched feature distributions. This ensures that we reduce the number of false matches. The face in the database with the largest number of matching points that agrees with the spatial distributions of the keypoints is considered as nearest face and is used for the classification of the new face.

## V. EXPERIMENTS

### A. Dataset

The frontal views of the mugshots are usually with neutral expression. Our mugshot dataset consists of 100 face images downloaded from **http://www.thesmokinggun.com/mugshots.** Some examples are shown in Fig. 2.

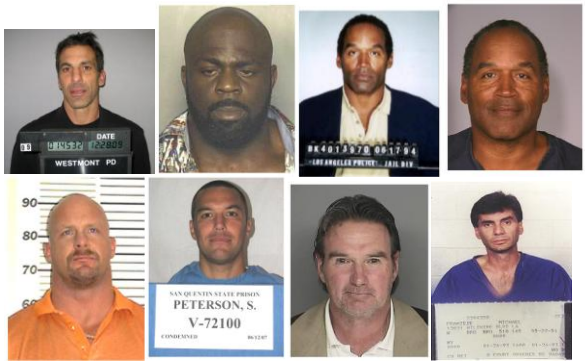

Figure 2. Frontal view of mugshots

Images are of different resolutions varies from 321x442 to 700x875. Only the face portion is cropped and used for evaluation. We have created 100 query face images from the 100 database images by performing various transformations to the database images. This is done by using the face transformation tool implemented by the University of St. Andrews available online in **http://morph.cs.st-andrews.ac.uk.** Original face image and its transformed versions are shown in Fig. 3. Further, images are converted to gray scale and resized to 300x300 pixels to assess the efficiency of the algorithms considered for comparison.

### B. Results

The aim of this face identification system is to measure similarity between query face image with all face images of database and retrieve the image which has got highest similarity i.e. nearest neighbour. Since query is created from one of the database images (source or original image), the system should retrieve that specific original face image. We have 100 manipulated faces as queries and 100 original face images of the criminals in the database.

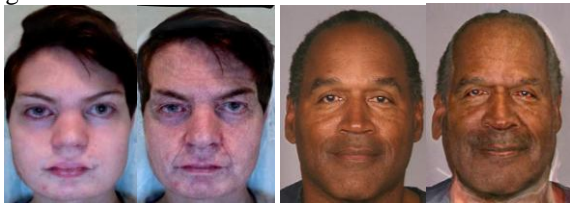

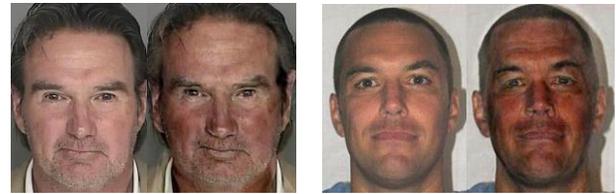

Figure 3. Original face images and its transformed versions

In order to evaluate performance of the system we input each query at a time. The identification rate is computed as follows;

$$Identification\ Rate = \frac{Number\ of\ times\ Correct\ Positives\ \text{Re}\,trieved}{Number\ of\ Queries} \times 100$$

Some resultant face images from both SIFT and PCA are shown in Fig. 4. Figure 4 shows two false positives and one correct positive retrieved using PCA and correct positives retrieved in all three cases using SIFT approach. The face identification rate is shown in Table 1. It is evident from Table 1 that SIFT performs better in face identification even under deliberate modifications.

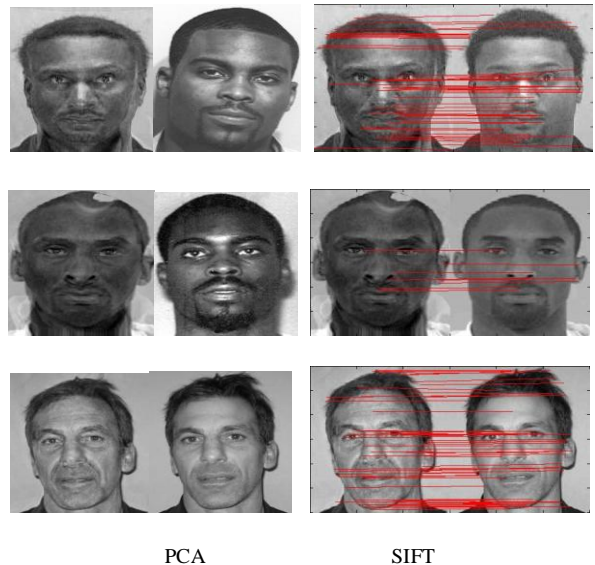

PCA             SIFT

Figure 4. Retrieved faces from PCA and SIFT approaches

TABLE I.    FACE IDENTIFICATION RATE

|                     | SIFT | PCA |
|---------------------|------|-----|
| Identification Rate | 92%  | 58% |

## VI. DISCUSSION

This paper presents a new approach for face identification from manipulated facial images based on SIFT features. The proposed approach is compared with eigenfaces and proved its superiority through experiments. In this paper, we concentrated only on mugshot identification. As an extension, we are investigating the use of SIFT features for retrieval of correct face with other forms of face representatives.

REFERENCES

[1] Robert G. Uhl, Jr., Niels da Vitoria Lobo, A Framework for Recognizing a Facial Image from a Police Sketch, IEEE Computer Society Conference on Computer Vision and Pattern Recognition, pp.586, 1996.

[2] Xiaogang Wang, Xiaoou, Face Photo-Sketch Synthesis and Recognition, IEEE Trans. On Pattern Analysis and Machine Intelligence, 31(11), pp. 1955-1967, Nov. 2009.

[3] Wolfgang Konen, Comparing Facial Line Drawings with Gray-Level Images: A Case Study on PHANTOMAS, Lecture Notes In Computer Science, Vol. 1112, Pages: 727 – 734, 1996.

[4] Yongsheng Gao, M.K.H. Leung, Siu Cheung Hui, and M.W. Tananda. Facial expression recognition from line-based caricatures, IEEE Transactions on Systems, Man, and Cybernetics, 33(3), pp. 407-412, 2003.

[5] Richa Singh, Mayank Vatsa, Afzel Noore, Sanjay K. Singh, Age Transformation for Improving Face Recognition Performance, Lecture Notes In Computer Science, pp. 576-583, 2007.

[6] Mohamed Aly, Face recognition using SIFT features
http://www.vision.caltech.edu/maala/research.php

[7] Han Yanbin, Yin Jianqin, Li Jinping, Human Face Feature Extraction and Recognition Base on SIFT, International Symposium on Computer Science and Computational Technology, vol. 1, pp.719-722, 2008.

[8] Lowe D.G., Distinctive Image Features from Scale-Invariant Keypoints International Journal of Computer Vision, 1(60), pp. 63-86, 2004.

[9] Matthew A. Turk, Alex P. Pentland, Face recognition using eigenfaces, In Proc. IEEE Computer Society Conference on Computer Vision and Pattern Recognition, 1991.

[10] H. R. Chennamma, Lalitha Rangarajan, Robust Near-Duplicate Image Matching for Digital Image Forensics, International Journal of Digital Crime and Forensics, 1(3). July 2009.